\documentclass[conference]{IEEEtran}

\usepackage{cite}
\usepackage{amsmath,amssymb,amsfonts}
\usepackage{graphicx}
\usepackage{textcomp}
\usepackage{xcolor}

\usepackage{times}
\usepackage{soul}
\usepackage[utf8]{inputenc}
\usepackage[small]{caption}

\usepackage{latexsym}
\usepackage{multirow}
\usepackage[noend]{algpseudocode}
\usepackage{url}
\definecolor{newcolor}{rgb}{.8,.349,.1}

\usepackage{amsthm}
\usepackage{subfigure}
\usepackage[]{todonotes}
\usepackage{centernot}
\usepackage[]{algorithm2e}

\newtheorem{definition}{Definition}

\DeclareMathOperator*{\argmax}{arg\,max}

\def\BibTeX{{\rm B\kern-.05em{\sc i\kern-.025em b}\kern-.08em
    T\kern-.1667em\lower.7ex\hbox{E}\kern-.125emX}}
\begin{document}

\title{Conceptual Domain Adaptation Using \\ Deep Learning}

\author{\IEEEauthorblockN{Behrang Mehrparvar}
\IEEEauthorblockA{\textit{Department of Computer Science} \\
\textit{University of Houston}\\
Houston, TX, USA \\
bmehrparvar@uh.edu}
\and
\IEEEauthorblockN{Ricardo Vilalta}
\IEEEauthorblockA{\textit{Department of Computer Science} \\
\textit{University of Houston}\\
Houston, TX, USA \\
rvilalta@uh.edu}
}

\maketitle

\begin{abstract}
Deep learning has recently been shown to be instrumental in the problem of domain adaptation, where the goal is to learn a model on a target domain using a similar --but not identical-- source domain.  The rationale for coupling both techniques is the possibility of extracting common concepts across domains. Considering (strictly) local representations, traditional deep learning assumes common concepts must be captured in the same hidden units. We contend that jointly training a model with source and target data using a single deep network is prone to failure when there is inherently lower-level representational discrepancy between the two domains; such discrepancy leads to a misalignment of corresponding concepts in separate hidden units. We introduce a search framework to correctly align high-level representations when training deep networks; such framework leads to the notion of conceptual --as opposed to representational-- domain adaptation.\footnote{This work has been submitted to IEEE for possible publication. Copyright may be transferred without notice, after which this version may no longer be accessible.}

\end{abstract}


\section{Introduction}
\label{sec:intro}

Many practical machine-learning applications, such as sentiment classification, spam filtering, and object recognition, require the repetitive building of new predictive models as fresh data becomes readily available. Assuming the classification of new data is laborious or costly, it is common to encounter a new target domain with a shortage of class labels, together with a previously analyzed source domain with an abundance of class labels. Under the assumption that both domains share the same feature (input) representation, one is tempted to use a model trained on the source domain and apply it on the target domain. The problem of domain adaptation emerges when source and target domain distributions differ; the discrepancy between the two domains precludes applying the source model on the target domain directly \cite{BenDavid10,Daume06,mansour2009domain,Shi12,Zhang13}. Based on certain assumptions, domain adaptation techniques have been proposed to alleviate such distributional discrepancy.

A popular technique in domain adaptation is to search for a new common feature space where both source and target distributions show high overlap. Deep learning has been recently used successfully in this scenario \cite{glorot_domain_2011,ganin_unsupervised_2014,ghifary_domain_2014}; the goal is to use a deep network architecture to transform low-level features into high-level representations. Features detected using deep networks have been shown to capture specific underlying factors of variation in the data, while being robust to other variations \cite{goodfellow2009measuring}.

In this paper we introduce the notion of \textit{conceptual domain adaptation} in which high-level concepts within source and target domains are identified and aligned in order to define a common feature space.  When source and target domains contain concepts with known semantic similarity, but marked difference in low-level representations, traditional domain adaptation techniques using deep learning fail to unify both domains. Conceptual domain adaptation, in contrast, focuses on the alignment of high-level concepts only, which provides the ability to solve a wider range of problems through imposing less constraints on the relation between the two domains.

The paper is organized as follows. Section~\ref{sec:background} gives background information on deep learning and de-noising auto-encoders. Section~\ref{sec:related} shows related work combining deep learning with domain adaptation. Our main methodology and fundamental ideas are described in Section~\ref{sec:conceptual}. Section~\ref{sec:joint} explains how jointly training a model on source and target data does not guarantee proper node alignment. Section~\ref{sec:proposed} shows our framework for aligning high-level representations, leading to the notion of conceptual domain adaptation. Experiments and results are described in Section~\ref{sec:results}. Finally, Section~\ref{sec:conclusions} gives our conclusions.

\section{Background Information}
\label{sec:background}

Deep learning networks iteratively learn multiple layers of intermediate non-linear data representations (i.e., data abstractions). Each layer contains a set of nodes that compute a non-linear combination of the output values of the nodes in the adjacent layer below. Although according to the Universal Approximation Theory \cite{csaji2001approximation}, a feed-forward neural network with only one hidden layer and sufficient hidden units is able to approximate any continuous function, deep architectures bring about added benefits, such as the ability to do feature re-use and feature abstraction. Re-using features not only yields a reduction in the number of computational nodes, it also reduces the number of parameters of the model, and thus the need for more samples. Furthermore, abstract features emerging from the network tend to show more resilience to data variations.

\subsection{Stacked De-Noising Auto-Encoders}\label{sec:SDAE}

An auto-encoder is a three layer neural network consisting of an input layer \(x\), middle layer \(h\), and output layer \(z\) \cite{bengio2009learning}. The network encodes data in the middle layer \(h\) and decodes data in the output layer \(z\). The output layer and hidden layers are composed of processing units (e.g., logistic sigmoid functions).

The goal of an auto-encoder is to learn the hidden layer \(h\) (i.e., the weights performing the coding step) by reconstructing the input on the output. This is achieved by minimizing a loss function (i.e.,  reconstruction error):
\(L(x,z)=||x-z||\). The minimization is performed using gradient descent by iteratively updating the weights of the encoder and the decoder. In de-noising auto-encoders, the network encodes a corrupted version of the input while trying to do the reconstruction. Accordingly, the network is forced to capture statistical dependencies between input features to filter out noise.

In order to capture complex feature abstractions, multiple layers of non-linear units are required. This is attainable using \textit{stacked} de-noising auto-encoders (SDAE)~\cite{vincent2010stacked}, essentially made of a layer-wise training of multiple auto-encoders. The input of each auto-encoder is composed of the hidden layer of the auto-encoder trained in the previous iteration. Here we refer to them as \textit{deep auto-encoders}. 

\section{Related Work}\label{sec:related}

The idea of transforming source and target data into a common space as an effective solution to the distribution discrepancy problem in domain adaptation has recently seen a surge of different techniques using deep learning architectures. As an example, \cite{glorot_domain_2011} (2011) proposed learning intermediate representations using stacked de-noising auto-encoders; the higher-level representation is learned using information from both source and target domains; the classifier is finally trained in the new space using source data only. Following a similar approach, \cite{chen_marginalized_2012} (2012) used marginalized stacked de-noising auto-encoders as an alternative architecture exhibiting lower computational costs, and better scalability on high-dimensional feature spaces. \cite{nguyen2015dash} (2015) proposed using a sparse and hierarchical network (DASH-N) for domain adaptation that is similarly trained  using source and target data jointly. \cite{ghifary_deep_2014} (2014a) proposed an alternative architecture that imposes sparse locally-connected weights in the bottom layer, in addition to the use of a sparsity regularizer.

Another research direction is to explore new cost functions during training. As an example, \cite{ganin_unsupervised_2014} (2014) proposed a deep network architecture to jointly learn a common representation space by minimizing reconstruction error and distribution discrepancy using a technique named gradient reversal layer. \cite{ghifary_domain_2014} (2014b) proposed an architecture that minimizes the maximum mean discrepancy between source and target distributions. Other work has focused on the use of regularizers during training. For example, \cite{ghifary_deep_2016} (2016) proposed a multi-task learning architecture for domain adaptation composed of an encoder, followed by a source class predictor, and target reconstruction; the target-reconstruction network works as a regularizer to prevent the source classifier from data overfitting. A different direction is to alleviate the distribution discrepancy between source and target by shifting domains. \cite{kan_bi-shifting_2015} (2015) proposed an architecture composed of one encoder followed by two decoders one for each domain; the encoder captures source and target into a common space, while each decoder is responsible for its corresponding domain. 

\section{Conceptual Domain Adaptation}\label{sec:conceptual}

A high-level concept in domain \(D\) can be captured by a pattern of output values at the top level of a deep network; it is an abstract entity that is assumed to carry a clear semantic meaning. For instance in the domain of hand-written digits, \textit{seven} is a high-level concept that carries the same meaning regardless of the writing form or style. 
More formally, we say that concept \(c_i^s\) in domain \(D^s\) (source domain) is \textit{correspondent} to concept \(c_j^t\) in domain \(D^t\) (target domain), \(c_i^s \leftrightarrow c_j^t\), if and only if they carry the same semantic meaning.

In this paper, we will refer to the representation $r$ of a concept $c$ as a binary vector corresponding to the output of the last hidden layer $h$ of a deep auto-encoder, after performing a layer-wise training of multiple auto-encoders (Section~\ref{sec:background}). Vector $r$ is obtained by applying a step function on each node on $h$. Specifically, assuming $h_j$ is the \(j\)th node in $h$, and $o(h_j)$ is the output of node $h_j$, then $r(h_j) = 1$ if $o(h_j) \geq 0.5$, and $0$ otherwise. 

Concepts can be represented in two main forms: using \textit{local} or \textit{distributed} representations. In local representations, activation of one hidden unit is necessary to represent the concept; in distributed representations, the concept is represented by an activation pattern over more than one hidden unit~\cite{wilson2001encyclopedia}. In higher layers of deep networks, representations tend to be more local, rather than distributed. In this paper we consider (strictly) local representations of high-level concepts defined as follows:

\begin{definition}[strictly local representation]
A representation is local~\cite{thorpe1998localized} if for each concept only one (output) unit is active (Figure~\ref{fig:local}).

\begin{equation}
\forall r \in c\ \ \exists h_j : r(h_j)=1, \forall h_k,j\neq k:r(h_k)=0
\end{equation}
where \(r \in c\) is one representation of concept \(c\) and \(r(h_j)\) is the \(j\)th hidden unit of representation \(r\).

In case of strictly local representations, activation of other units does not affect the representation of the concept~\cite{wilson2001encyclopedia}(Figure~\ref{fig:s_local}).

\begin{equation}
\forall r \in c\ \  \exists h_j : r(h_j)=1
\end{equation}

\begin{figure}[bth]
\vspace*{-6mm}
\centering    
\subfigure[local]{\label{fig:local}\includegraphics[width=0.20\textwidth]{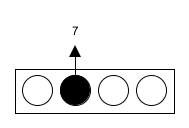}}
\subfigure[strictly local]{\label{fig:s_local}\includegraphics[width=0.20\textwidth]{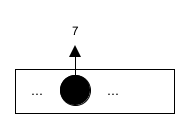}}
\vspace*{-2mm}
\caption{Unlike local representations (a) requiring other hidden units to be deactivated to represent a concept (e.g.,  concept seven), in strictly local representations (b), the activation of other hidden units does not affect the representation of such concept.}
\label{fig:local_rep}
\end{figure}

\end{definition}

\begin{definition}[Aligned local representations]\label{def:align}
Assuming (strictly) local representations, representation \(r^s\) of concept \(c^s\) in domain \(D^s\) is aligned to representation \(r^t\) of its correspondent concept \(c^t\) in domain \(D^t\), if and only if the activating unit of \(r^s\) is also activated in \(r^t\).

\begin{equation}
\forall c_i^s \in D^s, c_j^t \in D^t : c_i^s \leftrightarrow c_j^t \iff (\forall r^s \in c_i^s, r^t \in c_j^t : r^s = r^t)
\end{equation}
where
\begin{equation}\label{eq:aligned}
\forall r^s \in c^s, r^t \in c^t : r^s = r^t \iff ( r^s(h)=1 \iff  r^t(h)=1)
\end{equation}

\end{definition}

While many current approaches to domain adaptation are based on projecting source and target data into a new common space, our proposed approach extracts high-level concepts from each domain separately, followed by an alignment of correspondent concepts. As an illustration, consider the domain of hand-written digits and rotated hand-written digits. Here, concept seven in the first domain corresponds to the concept of rotated seven in the second domain, as they clearly carry the same semantic meaning. In order to perform domain adaptation, the two corresponding concepts must be aligned. Assuming a hierarchical representation of data (as is the case with deep neural networks), a domain-adaptation solution is considered \textit{conceptual} --as opposed to representational-- if the alignment between high-level correspondent concepts in the two domains does not rely on their low-level representations. Following up with the example described above, a representational approach to domain adaptation must rely on the lower-level pixel-wise relationship between seven and rotated seven to align the two concepts. In contrast, conceptual domain adaptation seeks to  align the two concepts while discarding information from low-level representational properties (e.g. pixel information).

We contend representational domain adaptation imposes a stringent limitation on the range of solvable problems, by focusing on those situations with low-level similarities between correspondent concepts across domains. By relaxing this limitation, conceptual domain adaptation leads to a less-constrained form of transfer knowledge across similar domains.

\section{The Problem Behind the Joint Training Approach}\label{sec:joint}

An important step in conceptual domain adaptation is to align correspondent concepts \(c^s\) and \(c^t\) such that the active hidden unit in representation  \(r^s\) of \(c^s\) is also active in  representation \(r^t\) of \(c^t\) (see definition \ref{def:align} for local alignment). This stands in stark contrast to previous approaches where no alignment takes place; an implicit assumption is made that the hidden unit capturing concept \(c^s\) is also able to simultaneously capture its correspondent concept \(c^t\).

To better understand the problem behind jointly training source and target domains without any form of concept alignment, consider that deep networks (e.g., deep auto-encoder) continuously update weights across the whole network, such that the formation of high-level concepts are dependent on the low-level representation of their constituent patterns. Now, in order to align correspondent concepts in the same hidden units using solely joint training, we would require that correspondent concepts exhibit lower-level representational similarities. This is rarely the case in real-world applications. A more common scenario occurs when correspondent concepts from two different domains exhibit inherently lower-level representational discrepancy (e.g., image of digit seven and image of rotated digit seven). In this case, the popular joint training approach for domain adaptation could capture correspondent concepts in two different hidden units, which would lead to an inevitable misalignment between the two semantically-identical concepts.

\subsection{The Effect of Concept Misalignment} 

A misalignment in the (local) representation of correspondent concepts between two domains (after joint-training) directly affects the performance of domain adaptation techniques. Consider concept \(c^s\) in the source domain represented by the activation of hidden unit \(h\) in local representation \(r^s\); correspondent concept \(c^t\) in the target domain is also trained using the same network and the two concepts are not locally aligned. The misalignment will result in deactivation of hidden unit \(h\) in \(r^t\) (the local representation of \(c^t\)).

\begin{figure}[!ht]
  \centering
      \includegraphics[width=0.4\textwidth]{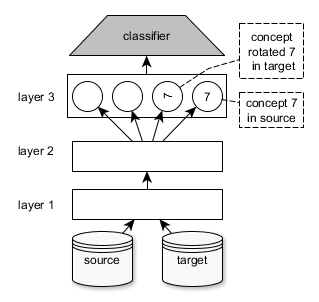}
  \caption{The problem of misalignment of source and target concepts using the joint-training approach.}
  \label{fig:problem}
\end{figure}

Figure~\ref{fig:problem} illustrates the misalignment problem using strictly local representations of high level concepts, under the joint-training approach.

We now introduce a search framework that provides a solution to the problem described above by adjusting 
high-level representations trained with deep learning (deep auto-encoders).

\section{Alignment and Adjustment}\label{sec:proposed}

Our methodology follows three main steps: (i) learning high-level concepts from 
source and target domains (independently) using deep auto-encoders, (ii) aligning correspondent concepts in source and target by adjusting their representations, and (iii) building a classifier on the aligned representations. Our main contribution lies on the second step, which we explain next. 

\subsection{Concept Alignment under the Mapping Matrix}\label{sec:mapping}

\begin{figure*}
\centering    
\subfigure[adjustment of target]{\label{fig:ref_mapping}\includegraphics[width=0.55\textwidth]{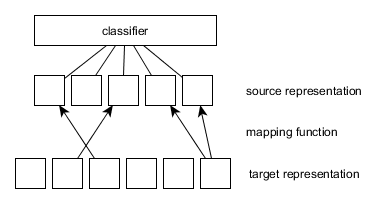}}
\subfigure[Mapping matrix and encoding]{\label{fig:matrix}\includegraphics[width=0.18\textwidth]{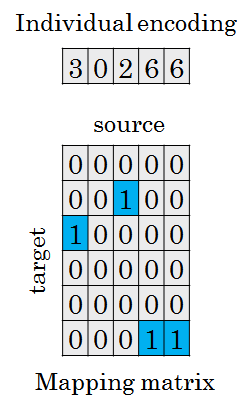}}
\vspace*{-2mm}
\caption{Alignment of target using adjustment approach.}
\vspace*{0mm}
\label{fig:refinement}
\end{figure*}

Our goal is to have correspondent concepts from target and source domains fall into the same hidden units along the upper  layer of a deep auto-encoder. As illustrated in Figure~\ref{fig:ref_mapping}, the target representation experiences an adjustment by using a mapping function that ensures concept correspondence with the source representation. Specifically, the target data is adjusted by defining a mapping function over hidden units (referred here as the nodes on the top hidden layer of a deep auto-encoder architecture). The mapping function for each hidden unit $h_k$ gives a new representation as follows:

\begin{equation}\label{eq:vs}
\begin{split}
r^n(h_j) = \sum_i{v_{ij}r^t(h_i)} \\
\textrm{s.t.}\ \ v_{ij} \in \{0,1\}, \ \forall j:\sum_i{v_{ij}}\leq 1
\end{split}
\end{equation}

\noindent
where $h_j$ is the node being adjusted, and $r^n(h_j)$ is the new representation for that node. Each weight $v_{ij}$ is restricted to a binary value, and at most one $v_{ij} = 1$ . In essence, the new representation \(r^n(h_j)\) will take the value of the old representation \(r^t(h_i)\) as specified by the position where $v_{ij} = 1$, or will take the value of $0$ if $\forall j\ \ v_{ij} = 0$ . As explained below, this can be seen as a mapping function intended to align the target and source representations.

The formulation above can be rephrased by defining a mapping matrix (Figure~\ref{fig:matrix}) with the number of rows and columns corresponding to the number of nodes on the target and source representations respectively. Similar to equation~\ref{eq:vs}, activation of each element in the mapping matrix corresponds to a mapping from the specified target unit to the corresponding source unit. Each column can only be activated by either one, or none of the units in the target representation (rows in the matrix).

Using mapping matrix \(M^*\), the aligned target representation can be computed through a linear transformation:

\begin{equation}
T_{\textrm{new}}=T\times M^* \mid M^*_{i,j}\in \{0,1\}, \forall j:\sum_i{M^*_{i,j}}\leq 1
\end{equation}

\noindent
where $T_{\rm new}$ and $T$ are the new and original training samples (target domain). The intuition behind this type of adjustment is based on the assumption that the (separate) training of the source and target deep auto-encoders had already been able to extract meaningful concepts from the data in both domains. As a result, an adjustment as proposed above should properly align correspondent high-level concepts.

At this point, the main challenge is to find an optimal mapping matrix \(M^*\) as follows:

\begin{equation}
M^*=\argmax_{M} G(S,T;M)
\end{equation}

\noindent
where $S$ and $T$ are the source and target samples, and $G(S, T; M)$ quantifies the goodness of matrix $M$.

\subsection{Search for Matrix $M^*$}\label{sec:search}

Finding matrix $M^*$ requires exploring a space of possible solutions, and a metric $G(S, T; M)$ to quantify the quality of each solution. The total number of possible solutions (combinations) for a $p \times q$ mapping matrix is \(p^q\). To handle such large space, our work employs genetic algorithms~\cite{eiben2003introduction}. Each mapping matrix \(M\) is encoded as an offspring $E(M)$ through a vector of integers \(V\) as follows:

\begin{equation}
E(M)=V\Rightarrow (\forall i:V(i)=j\iff M(j,i)=1)
\end{equation}

Figure~\ref{fig:matrix} illustrates the mapping matrix and the corresponding encoding of an individual solution. We use three main operations: elite selection, crossover and mutation, to generate the next population at each new iteration. The fitness value is obtained by training a $k$-NN classifier on the (high-level representation) of the source data and testing the performance of such classifier on the adjusted target data. We take accuracy as the fitness value. Pseudo-code to compute the fitness value of matrix \(M\) is shown in Algorithm~1.

\begin{algorithm}{}{\textbf{Algorithm 1: Score of matrix M}\\}
\SetAlgoLined
\KwData{source data \(S\), target data \(T\), matrix \(M\)}
\KwResult{score of mapping matrix \(M\)}
\(T_{\rm new}=T\times M\)\;
\(\theta^*= \textrm{train-kNN}(S)\)\;
\(\textrm{accuracy} = \textrm{classify}(T_{new},\theta^*)\)\;
return \(\textrm{accuracy}\)\;
\label{alg:score}
\vspace*{-5mm}
\end{algorithm}

\section{Experimental Settings}
\label{sec:experiments}

Our experiments used a stacked de-noising auto-encoder~\cite{vincent2010stacked} architecture for model training, and a modified version of Matlab's implementation of the deep network\footnote{https://github.com/rasmusbergpalm/DeepLearnToolbox}. The model comprised nine layers of de-noising auto-encoders, each using batch gradient descent for optimization. Training stopped if no improvement was achieved for the last \(20\) iterations, or if the number of iterations exceeded a threshold (\(500\) iterations). The learning rate $\epsilon_t$ was defined following the method described by \cite{bengio2012practical}, as follows:

\begin{equation}
\epsilon_t=\frac{\epsilon_0\tau}{\max(t,\tau)}
\end{equation}
where \(\tau\) is the minimum number of iterations to reduce the learning rate (set to \(\tau = 20\) iterations). We adopted a layer-wise search for hyper-parameters; specifically, we performed a grid search on sets of hyper-parameter values and opted for the best setting. The size of the hidden layer was chosen from \(\{L, 2L/3, L/2, L/5\}\) where \(L\) is the size of the previous layer. Similarly, the range of learning-rate values and corruption level hyper-parameters where chosen from the sets \(\{10^{-3} , 10^{-2}, 10^{-1}, 1\}\) and \(\{0.0, 0.3, 0.5\}\) respectively. 

Regarding the genetic algorithm, at each iteration we kept \(20\%\) of the population as elite instances, the remaining \(80\%\) was generated using crossover and mutation. The size of the population was set to \(100\) instances. The algorithm stopped when the best score did not improve for the last \(200\) iterations. The nearest-neighbor classifier was set to \(k=1\) and adopted \(L_1\) distance. 

We applied domain adaptation to the digit recognition task and used the following datasets:  MNIST~\cite{lecun1998gradient}\footnote{http://yann.lecun.com/exdb/mnist/}, USPS~\footnote{http://statweb.stanford.edu/~tibs/ElemStatLearn/data.html} and rotated USPS. The datasets where processed following the standard format of \(16\times16\) grayscale images.

\section{Results and Analysis}\label{sec:results}

\subsection{The Role of Adjustment}\label{sec:role_refinement}

\begin{figure}[!ht]
  \centering
      \includegraphics[width=0.35\textwidth]{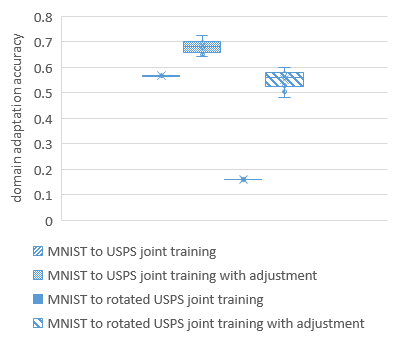}
   \vspace*{-2mm}
  \caption{Improvement in accuracy with domain-adaptation and search-based adjustment}
  \label{fig:exp_adjustment}
\end{figure}

For the first batch of experiments, we tested the adjustment approach on two domain-adaptation scenarios including MNIST to USPS and MNIST to rotated USPS. Based on accuracy performance, we limited the architecture to 5 layers, with each layer being 2/3 of the size of the previous layer. Figure~\ref{fig:exp_adjustment} shows the improvement gained with our proposed adjustment during domain adaptation on each scenario.

In the MNIST to USPS scenario, correspondent concepts between the two domains have stronger representational similarities compared to the other scenario (MNIST to rotated USPS); the proposed approach shows only a small performance gain. On the MNIST to rotated USPS scenario, the adjustment approach displays a significant improvement due to the high degree of low-level representational discrepancy. Overall, the proposed search-based framework shows performance gain, regardless of the presence --or lack of-- low-level representational discrepancy.

\subsection{The Role of Depth in the Auto-encoder}\label{sec:role_depth}

\begin{figure}[!ht]
  \centering
  \hspace*{0mm}
      \includegraphics[width=0.27\textwidth]{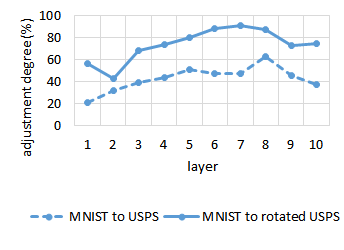}
      \hspace*{0mm}
      \vspace*{0mm}
      \includegraphics[width=0.27\textwidth]{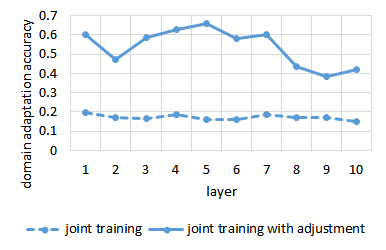}
   \vspace*{-2mm}
   \caption{(top) Adjustment degree as a function of the number of layers in the deep auto-encoder. (bottom) The effect of depth on domain adaptation performance for MNIST to rotated USPS scenario.}
  \label{fig:role_depth}
\end{figure}

In order to assess the effectiveness of the proposed mapping with respect to the number of hidden layers in the deep auto-encoder, we captured the deviation of $M^*$ from the identity matrix $I_n$ (where we assume $M^*$ and $I_n$ are square matrices). In the extreme case where $M^* = I_n$, the mapping is direct: concepts are represented by the same hidden unit. Figure~\ref{fig:role_depth}  (top) shows the adjustment degree for each number of of layers in a deep auto-encoder, on the two domain-adaptation scenarios described above. Adjustment degree is the percentage of hidden nodes in the new representation that had to be changed (adjusted) to be correctly mapped from the corresponding hidden unit in the old target representation.

Results show how lower layers lead to an increase of direct mappings (more identity mappings). This is to be expected: low-level features contain more "representational" rather than "conceptual" relationships; here representational alignments adopted by traditional joint-training approaches suffice to achieve good results. 

Figure~\ref{fig:role_depth} (bottom) compares accuracy versus depth (number of layers). We observe that when our proposed adjustment is invoked, accuracy exhibits significant variation, and there is usually an optimal value that maximizes performance. However, the performance seen when using solely joint training does not show much variation with depth. The same behavior is observed in the MNIST to USPS scenario. We conclude that conceptual domain adaptation --as compared to joint-training-- justifies a search for a performance maximum, with depth as the control parameter. Figure~\ref{fig:role_depth} (bottom) shows a decrease in accuracy using conceptual domain adaptation past layer 5. This can be explained by the increase in adjustment degree (Figure~\ref{fig:role_depth}) at high layers, where more adjustments are required to align  correspondent units. This is also an effect of our 2/3 rule in the design of the network architecture: while reducing the size of the network, more information is lost and performance degradation accrues.

\subsection{The Role of Jointly Learning New Concepts}\label{sec:role_joint}

\begin{figure}
\vspace*{-2mm}
\centering     
\subfigure[MNIST to USPS]{\label{fig:exp_joint_USPS}\includegraphics[width=0.23\textwidth]{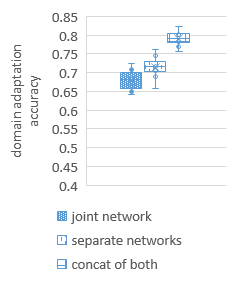}}
\subfigure[MNIST to rotated USPS]{\label{fig:exp_joint_USPS_R}\includegraphics[width=0.23\textwidth]{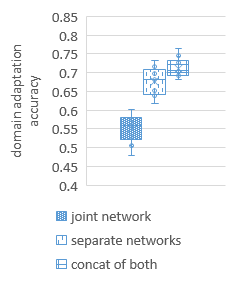}}
\caption{The effect of joint network versus separate networks training}
\label{fig:exp_joint}
\vspace*{-3mm}
\end{figure}

We now compare two approaches to test different learning strategies: (i) jointly learning the source and target concepts in the same network, (ii) separately learning the source and target concepts in different networks, and a third case (iii) where the representation of each data point is constructed by concatenating the representation obtained from the previous two approaches. For each case, the adjustment was performed by constructing the mapping matrix. The size of the mapping matrix is dependent on the size of the network used for training the source and target data in each case as follows:

In case (i), we follow the same approach as previous experiments, where we search for mapping matrix $M^*$; we assume the same number $n$ of rows and columns (corresponding to the number of hidden units in the highest layer of the network). In case (ii), the size of $M^*$ is \(p\times q\), where \(p\) and \(q\) correspond to the number of nodes along the highest layer of the deep auto-encoders trained with target and source datasets respectively (\(p\) and \(q\) may differ). Finally, in case (iii), $M^*$ is an \((n+p)\times (n+q)\) matrix initialized with four sub-matrices as follows:

\begin{equation}
M=
\begin{bmatrix}
	J && 0 \\
    0 && S
\end{bmatrix}
\end{equation}
where \(J\) is the a sub-matrix corresponding to the joint training of source and target, and  \(S\) corresponds to the case where source and target are learned separately. The \(J\) and \(S\) matrices are initialized as a diagonal matrix and random matrix respectively. A diagonal mapping matrix is one where only the diagonal elements are activated, corresponding to a direct mapping from target and source hidden units. A random matrix is one where the activation of elements are randomly distributed; each source hidden unit is randomly mapped to one or none of the target units.  Figure~\ref{fig:exp_joint} compares accuracy among all three approaches.

Results show that separately training source and target improves accuracy significantly, compared to using a single-joint network. The reason can be traced to the inability of the joint network to capture correspondent concepts for each domain separately when there is low representational similarity. We conclude that a single joint network architecture is unable to map correspondent concepts under low representational relationships. In contrast, training a separate network for each domain facilitates capturing correspondent concepts, since there is no interference during learning. Results also show it is possible for a joint network to capture unique concepts common to both domains (under representational similarities). As illustrated in Figure~\ref{fig:exp_joint}, a reasonable approach is to use the concatenation of both representations to achieve high accuracy performance. 

\begin{figure}
\centering     
\subfigure[MNIST to Braille]{\label{fig:exp_Braille}\includegraphics[width=0.23\textwidth]{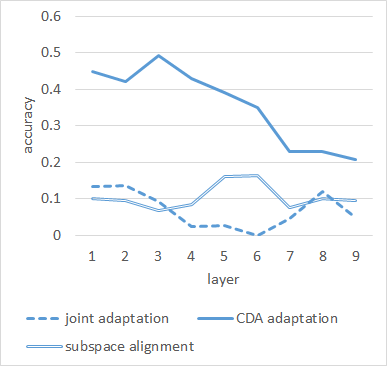}}
\subfigure[MNIST to SVHN]{\label{fig:exp_SVHN}\includegraphics[width=0.23\textwidth]{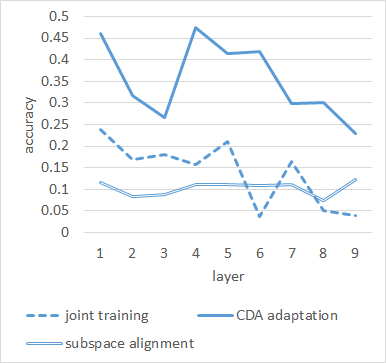}}
\caption{A comparison of our approach to:  no-adaptation, joint training, and subspace alignment.}
\label{fig:exp_more}
\vspace*{-4mm}
\end{figure}

Apart from semantic similarity, MNIST and (rotated) USPS have low-level similarities. To add more complexity to the adaptation process, we have experimented with the braille ~\footnote{Images of four dots colored black or white based on each digit.} dataset and the street-view house numbers (SVHN) datasets~\footnote{http://ufldl.stanford.edu/housenumbers/}. We also compare our approach with an additional domain-adaptation method known as subspace alignment~\cite{fernando2014subspace}. As shown in Figure~\ref{fig:exp_Braille} and Figure~\ref{fig:exp_SVHN}, conceptual domain adaptation outperforms joint training and subspace alignment in both scenarios. 

\vspace*{0mm}
\section{Conclusions}\label{sec:conclusions}
\vspace*{0mm}

This paper describes an approach to domain adaptation that employs deep learning to extract high-level concepts from source and target domains, while relaxing the alignment dependency on lower-level representations of correspondent concepts. The proposed alignment is based on adjusting the final (high-level) representation of the target data by matching the corresponding representation on the source data.

Experimental results show that our approach brings significant gains in accuracy, particularly under scenarios with high discrepancy in low-level representations. Increasing the depth of the network (i.e., of the deep auto-encoder) leads to more adjustments in order to align correspondent concepts. An abundance of representational discrepancy leads to more adjustments. Finally, we show that a combined approach that concatenates the representations of both the joint network and each of the domain networks yields best results on both experimental settings.


\bibliographystyle{IEEEtran}
\bibliography{DAviaDL}

\begin{thebibliography}{10}
\providecommand{\url}[1]{#1}
\csname url@samestyle\endcsname
\providecommand{\newblock}{\relax}
\providecommand{\bibinfo}[2]{#2}
\providecommand{\BIBentrySTDinterwordspacing}{\spaceskip=0pt\relax}
\providecommand{\BIBentryALTinterwordstretchfactor}{4}
\providecommand{\BIBentryALTinterwordspacing}{\spaceskip=\fontdimen2\font plus
\BIBentryALTinterwordstretchfactor\fontdimen3\font minus
  \fontdimen4\font\relax}
\providecommand{\BIBforeignlanguage}[2]{{%
\expandafter\ifx\csname l@#1\endcsname\relax
\typeout{** WARNING: IEEEtran.bst: No hyphenation pattern has been}%
\typeout{** loaded for the language `#1'. Using the pattern for}%
\typeout{** the default language instead.}%
\else
\language=\csname l@#1\endcsname
\fi
#2}}
\providecommand{\BIBdecl}{\relax}
\BIBdecl

\bibitem{BenDavid10}
S.~BenDavid, J.~Blitzer, K.~Crammer, A.~Kulesza, F.~Pereira, and J.~W. Vaughan,
  ``A theory of learning from different domains,'' \emph{Machine Learning},
  vol.~79, no.~1, pp. 151--175, 2010.

\bibitem{Daume06}
H.~D. III and D.~Marcu, ``Domain adaptation for statistical classifiers,''
  \emph{Journal of Artificial Intelligence Research}, vol.~26, pp. 101--126,
  2006.

\bibitem{mansour2009domain}
Y.~Mansour, M.~Mohri, and A.~Rostamizadeh, ``Domain adaptation: Learning bounds
  and algorithms,'' \emph{arXiv:0902.3430}, 2009.

\bibitem{Shi12}
Y.~Shi and F.~Sha, ``Information-theoretical learning of discriminative
  clusters for unsupervised domain adaptation,'' in \emph{Proceedings of the
  29th International Conference on Machine Learning}, ser. ICML'12.\hskip 1em
  plus 0.5em minus 0.4em\relax Omnipress, 2012, pp. 1275--1282.

\bibitem{Zhang13}
K.~Zhang, B.~Sch{\"o}lkopf, K.~Muandet, and Z.~Wang, ``Domain adaptation under
  target and conditional shift,'' in \emph{Proceedings of the 30th
  International Conference on Machine Learning}, 2013, p. 819–827.

\bibitem{glorot_domain_2011}
X.~Glorot, A.~Bordes, and Y.~Bengio, ``Domain adaptation for large-scale
  sentiment classification: A deep learning approach,'' in \emph{International
  Conference on Machine Learning}, 2011, pp. 513--520.

\bibitem{ganin_unsupervised_2014}
Y.~Ganin and V.~Lempitsky, ``Unsupervised domain adaptation by
  backpropagation,'' \emph{arXiv preprint arXiv:1409.7495}, 2014.

\bibitem{ghifary_domain_2014}
M.~Ghifary, W.~B. Kleijn, and M.~Zhang, ``Domain adaptive neural networks for
  object recognition,'' in \emph{Pacific Rim International Conference on
  Artificial Intelligence}.\hskip 1em plus 0.5em minus 0.4em\relax Springer,
  2014b, pp. 898--904.

\bibitem{goodfellow2009measuring}
I.~Goodfellow, H.~Lee, Q.~V. Le, A.~Saxe, and A.~Y. Ng, ``Measuring invariances
  in deep networks,'' in \emph{Advances in neural information processing
  systems}, 2009, pp. 646--654.

\bibitem{csaji2001approximation}
B.~C. Cs{\'a}ji, ``Approximation with artificial neural networks,''
  \emph{Faculty of Sciences, Etvs Lornd University, Hungary}, vol.~24, p.~48,
  2001.

\bibitem{bengio2009learning}
Y.~Bengio, ``Learning deep architectures for ai,'' \emph{Foundations and Trends
  in Machine Learning}, vol.~2, no.~1, pp. 1--127, 2009.

\bibitem{vincent2010stacked}
P.~Vincent, H.~Larochelle, I.~Lajoie, Y.~Bengio, and P.-A. Manzagol, ``Stacked
  denoising autoencoders: Learning useful representations in a deep network
  with a local denoising criterion,'' \emph{JMLR}, vol.~11, no. Dec, pp.
  3371--3408, 2010.

\bibitem{chen_marginalized_2012}
M.~Chen, Z.~Xu, K.~Weinberger, and F.~Sha, ``Marginalized denoising
  autoencoders for domain adaptation,'' \emph{arXiv:1206.4683}, 2012.

\bibitem{nguyen2015dash}
H.~V. Nguyen, H.~T. Ho, V.~M. Patel, and R.~Chellappa, ``Dash-n: Joint
  hierarchical domain adaptation and feature learning,'' \emph{IEEE
  Transactions on Image Processing}, vol.~24, no.~12, pp. 5479--5491, 2015.

\bibitem{ghifary_deep_2014}
M.~Ghifary, W.~B. Kleijn, and M.~Zhang, ``Deep hybrid networks with good
  out-of-sample object recognition,'' in \emph{IEEE International Conference on
  Acoustics, Speech and Signal Processing}.\hskip 1em plus 0.5em minus
  0.4em\relax IEEE, 2014a, pp. 5437--5441.

\bibitem{ghifary_deep_2016}
M.~Ghifary, W.~B. Kleijn, M.~Zhang, D.~Balduzzi, and W.~Li, ``Deep
  reconstruction-classification networks for unsupervised domain adaptation,''
  in \emph{European Conference on Computer Vision}.\hskip 1em plus 0.5em minus
  0.4em\relax Springer, 2016, pp. 597--613.

\bibitem{kan_bi-shifting_2015}
M.~Kan, S.~Shan, and X.~Chen, ``Bi-shifting auto-encoder for unsupervised
  domain adaptation,'' in \emph{IEEE International Conference on Computer
  Vision}, 2015, pp. 3846--3854.

\bibitem{wilson2001encyclopedia}
R.~A. Wilson and F.~C. Keil, \emph{The MIT encyclopedia of the cognitive
  sciences}.\hskip 1em plus 0.5em minus 0.4em\relax MIT press, 2001.

\bibitem{thorpe1998localized}
S.~Thorpe, ``Localized versus distributed representations,'' in \emph{The
  handbook of brain theory and neural networks}.\hskip 1em plus 0.5em minus
  0.4em\relax MIT Press, 1998, pp. 549--552.

\bibitem{eiben2003introduction}
A.~E. Eiben and J.~E. Smith, \emph{Introduction to evolutionary
  computing}.\hskip 1em plus 0.5em minus 0.4em\relax Springer, 2003, vol.~53.

\bibitem{bengio2012practical}
Y.~Bengio, ``Practical recommendations for gradient-based training of deep
  architectures,'' in \emph{Neural Networks: Tricks of the Trade}.\hskip 1em
  plus 0.5em minus 0.4em\relax Springer, 2012, pp. 437--478.

\bibitem{lecun1998gradient}
Y.~LeCun, L.~Bottou, Y.~Bengio, and P.~Haffner, ``Gradient-based learning
  applied to document recognition,'' \emph{Proceedings of the IEEE}, vol.~86,
  no.~11, pp. 2278--2324, 1998.

\bibitem{fernando2014subspace}
B.~Fernando, A.~Habrard, M.~Sebban, and T.~Tuytelaars, ``Subspace alignment for
  domain adaptation,'' \emph{arXiv preprint arXiv:1409.5241}, 2014.

\end{thebibliography}

\end{document}